\documentclass[journal]{IEEEtran}

\usepackage{graphicx}
\usepackage{amsmath}
\usepackage{amssymb}
\DeclareMathOperator*{\argmax}{argmax}
\DeclareMathOperator*{\argmin}{argmin}
\usepackage{algorithm}
\usepackage{algorithmic}
\usepackage{multirow}
\usepackage{xcolor}
\usepackage{colortbl}
\usepackage{dsfont}
\usepackage{framed}
\usepackage{diagbox}
\usepackage{float}

\usepackage{fancyhdr}

\begin{document}
\title{Driver Identification through Stochastic Multi-State Car-Following Modeling}

\author{Donghao~Xu,
        Zhezhang~Ding,
        Chenfeng~Tu,
        Huijing~Zhao,
        Mathieu~Moze,
        Fran\c{c}ois~Aioun,
	and~Franck~Guillemard
\thanks{This work is partially supported by Groupe PSA's OpenLab program (Multimodal Perception and Reasoning for Intelligent Vehicles) and the NSFC
Grants 61973004. D.Xu and Z.Ding are both the first authors of this paper. D. Xu, Z. Ding, C. Tu and H. Zhao are with the Key Lab of Machine Perception (MOE), Peking University, Beijing, China; M. Moze, F. Aioun and F. Guillemard are with Groupe PSA, Velizy, France. Contact: H. Zhao, zhaohj@pku.edu.cn.}
}

\markboth{IEEE TRANSACTIONS ON INTELLIGENT TRANSPORTATION SYSTEMS,~Vol.~?, No.~?, May~2020}%
{Shell \MakeLowercase{\textit{et al.}}: IEEE TRANSACTIONS ON INTELLIGENT TRANSPORTATION SYSTEMS}

\maketitle

\begin{abstract}
Intra-driver and inter-driver heterogeneity has been confirmed to exist in human driving behaviors by many studies. In this study, a joint model of the two types of heterogeneity in car-following behavior is proposed as an approach of driver profiling and identification. It is assumed that all drivers share a pool of driver states; under each state a car-following data sequence obeys a specific probability distribution in feature space; each driver has his/her own probability distribution over the states, called driver profile, which characterize the intra-driver heterogeneity, while the difference between the driver profile of different drivers depict the inter-driver heterogeneity. Thus, the driver profile can be used to distinguish a driver from others. Based on the assumption, a stochastic car-following model is proposed to take both intra-driver and inter-driver heterogeneity into consideration, and a method is proposed to jointly learn parameters in behavioral feature extractor, driver states and driver profiles. Experiments demonstrate the performance of the proposed method in driver identification on naturalistic car-following data: accuracy of 82.3\% is achieved in an 8-driver experiment using 10 car-following sequences of duration 15 seconds for online inference. The potential of fast registration of new drivers are demonstrated and discussed.
\end{abstract}


%
\IEEEpeerreviewmaketitle

\thispagestyle{fancy}
\fancyhead{}
\lhead{}
\lfoot{\small\copyright 2020 IEEE. Personal use of this material is permitted. Permission from IEEE must be obtained for all other uses, in any current or future media, including reprinting/republishing this material for advertising or promotional purposes, creating new collective works, for resale or redistribution to servers or lists, or reuse of any copyrighted component of this work in other works.}
\cfoot{}
\rfoot{}

\section{Introduction}

\IEEEPARstart{T}{he} driving style of each driver is different. Heterogeneities exist in the way a driver operates on steering wheel, gas and break pedals etc. in performing certain behaviors, which turns out different driving styles correlating with road and scene vehicles. Treating such heterogeneities as a kind of signature, many researches have been conducted in classifying driving style or understanding driver state of such as sporty, normal or comfortable, evaluating driving skill or recognizing identity of the driver \cite{14}\cite{qu2015switching-based}, which are crucial for the potential applications such as situation-based or personalized assistance.

On the other hand, car following is an essential component of a driver's behavior, where heterogeneity has been studied as an important facet that is a consequence of human factors in this driving process \cite{9}. The level of heterogeneity in the car-following behaviors of different drivers is substantial \cite{2} as well as of vehicle/driver combinations \cite{3}, which is called inter-driver heterogeneity. The internal stochasticity of an individual driver, called intra-driver heterogeneity, is another rational cause for the randomness of car-following behaviors \cite{10}.

This research studies driver identification by modeling both intra-driver and inter-driver heterogeneities in car following behaviors. Inspired by time series classification approaches that are based on the bag-of-words encoding scheme, intra-driver heterogeneity is modeled in a stochastic multi-state procedure; a code book containing the potential states of all drivers, and driver profiles recording multi-state distributions of each individual are learnt, which are used to infer the driver identity given car-following data sequences.

We claim that this work has the following novel contributions: 1) a method is developed to learn a stochastic model, i.e. a specific mixture of Gaussian driver states for each driver, that represents both intra- and inter-driver heterogeneities in car following behaviors; 2) a driver identification method is developed with it by characterizing individual differences and changes in an individual's behavior as a probability distribution over states; 3) experiments are conducted using the naturalistic driving data that are collected on the motorways in Beijing, and results show that online inference of driver's identity has an increased accuracy with more data samples, where 82.3\% is achieved with 10 car-following samples, 15 seconds per sample, in an 8-driver experiment.

Comparing with other literature works, our driver identification method has the following features: 1) it is a scene-aware method, where a drivers' response to scene objects, i.e., leading vehicle in this work, is addressed, which is an essential feature in describing a driver's state and the meticulous behavior in traffic, 2) it models car following behaviors using the Gaussian mixture model (GMM), where Gaussian driver states are learnt, and drivers are profiled using different mixture weights to represent the intra- and inter-driver heterogeneities of the behavior, 3) it is an online method that infers driver identity by using short sequences of car following data. Furthermore, an underlying assumption of this work is that although intra- and inter-driver heterogeneities exist, different drivers' behavior in short time segments share many commonalities, and all drivers can be modeled with a limited number of driver states. Therefore, the proposed method has a potential of registering a new driver without re-training of the Gaussian driver states, which is demonstrated by a preliminary experiment in this work and will be elaborated in future.

The remainder of the paper is organized as follows. Section II gives literature review, section III describes the proposed approach of driver identification through stochastic multi-state car following modeling. Experimental results are presented in section IV, followed by conclusion and future work in section V.

\section{Literature Review}

\subsection{Heterogeneity in Car-Following Behaviors}

Car-following is an essential component of a driver's behavior that has been studied for decades due to its important significance in modern transportation systems. Many car following models have been developed \cite{1}, where the process of a driver's behavior is generally described as a transformation from some perceived information about the driving situation, such as the speed and distance of a leading vehicle relative the ego (i.e. the follower), and the ego's speed, to control actions for acceleration or deceleration. Recent researches are addressed focusing more on modeling the heterogeneity in car following behaviors. \cite{2} quantified the extent of heterogeneity by analyzing a trajectory data collected in real-world traffic, which turned out that different drivers react differently to the stimuli from a leading car, \cite{3} extended the work by relating such observed heterogeneity to vehicle types in the composition of follower-leader pairs, \cite{4} calibrated a full set of random coefficients that account for the heterogeneity across drivers, and \cite{12} proposed a stochastic framework that takes into account both individual and general driving characteristics as one aggregate model. Intra-driver heterogeneity has been studied as well, where \cite{5} divided the procedure in different regimes and modeled acceleration control at each particular situation, \cite{8} incorporates stochastic Markov regime switching model to address the driving features at different regimes, \cite{6} developed a hierarchical Bayesian model with time-varying parameters to account for the gradual changes in car-following behaviors, \cite{taylor2015method} use dynamic time warping (DTW) to calibrate time-varying response times and critical jam spacing of a car following model, which are further used to analyze the intra-driver heterogeneity and situation-dependent behavior within a trip, and \cite{7} segmented a continuous stream into clusters by evaluating similarities on driving features. In addition, heterogeneity related with roadway categories has also been reported \cite{11}.

Such inter- and intra-driver heterogeneity in car following behavior can be used as a signature prompting a drivers identity or state.

\subsection{Driver Profiling, Identification of Behavior Characteristics}

\begin{figure*}[tb]
  \begin{center}
    \includegraphics[keepaspectratio=true,width=1.0\linewidth]{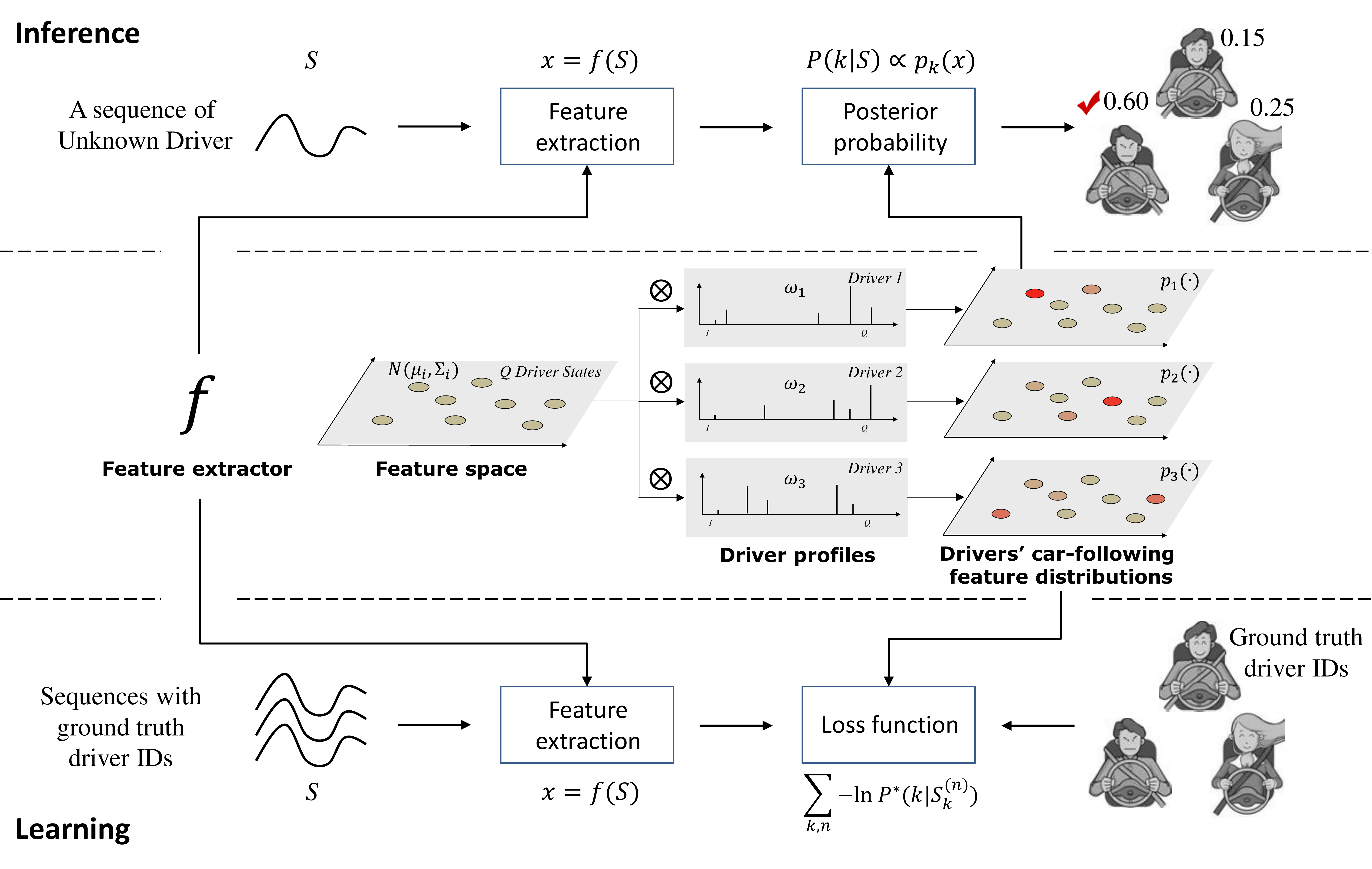}
  \end{center}
\caption{Framework of the proposed approach.}
\label{framework}
\end{figure*}

Driver profiling is among the most important applications of driver behavior studies \cite{14}, which addresses heterogeneities in driving behaviors to characterize particular driving styles, level driving skills and recognize driver identities.
Although these works can all be formulated as classification problems, they are much different. The studies of driving style or skill fall into the same category \cite{qu2015switching-based}, where the class definitions are subjective and ambiguous. To leverage the problem, only a small number of very rough groups are usually defined, e.g. normal/aggressive \cite{17}\cite{18}, aggressive/cautious/super cautious \cite{19}, sporty/normal/comfortable \cite{16}, normal/vague/aggressive \cite{wang2017driving}, low-/moderate-/high-risk \cite{li2017estimation} or cautious/average/expert/reckless \cite{20} driving styles, and low/high \cite{15} or experienced/average driving skills. Rather than relying on subjectively defined classes, \cite{wang2019driving} developed an unsupervised learning method use Bayesian nonparametric approach. Long driving sequences are first quantized to primitive driving patterns, the distributions of which are used to analyze individual driving styles and the similarity between drivers with an entropy index. In these studies, ground truth is usually difficult to be obtained, performance evaluation of the algorithms are still open questions.
The problem of driver identification is different as driver identity is definite and unique. Many methods have been developed by identifying drivers using the signals such as gear, gas pedal, engine, steering wheel, vehicle speed, yaw rate etc.\cite{enev2016automobile}\cite{martinez2016driver}\cite{jafarnejad2017towards}\cite{marchegiani2018long-term}. A challenge is faced when large number of classes need to be addressed and/or data resources are limited. To leverage this problem, \cite{21} assumes that a typical family has two to three drivers for a single car, hence formulates driver differentiation as a 2-class or 3-class problem and inferred by using only inertial sensors. However most of the works focus on characterizing a driver's general profile of its control operations, without correlating them with certain maneuvers nor driving scene, and the intra- heterogeneity of a driver's behavior is usually neglected. \cite{tanaka2019large-scale} addresses more than 10,000 drivers, where the data such as speed, acceleration, time and location are used that is collected by GPS sensors equipped with smart phones. The method model location and time correlated behaviors using long-term driving data, but may have difficulties in on-line inference for driving assistance. \cite{22} analyzes drivers operations during a turn, where a sequence of behaviors are needed to slow down, turn on the blinker, rotate steering wheel, accelerate etc., which obtained average classification accuracies of 76.9\% and 50.1\% in two and five drivers experiments respectively. \cite{23} is the only work in literature that identify drivers by modeling car following patterns on such as headway distance and relative velocity. However, only short pieces of data segments (3-5 min) are studied. This research studies driver identification using vehicle's driving sequences in car following, which has a special focus on answering the question: whether a driver can be identified by modeling the heterogeneous control operations of a certain maneuver that are correlative with scene objects, i.e., the leading vehicle in this work.

\subsection{Time-Series Classification}

Car-following data are measured sequences over time. Diver identification based on car-following data can be formulated generally as a time series classification (TSC) problem, which has broad applications in speech recognition, signature verification, financial analysis etc. and a plethora of methods have been developed in literature \cite{Bagnall2017The}\cite{fawaz2019deep} that can be divided into three groups. Instance-based approaches classify time series by comparing the sequential values of either the whole or sub sequences, e.g. shapelet \cite{hills2014classification}, which are usually exploited to solve the problems when class membership can be define by the similarity in shape, and DTW (dynamic time warping) \cite{jeong2011weighted} can be used to mitigate shape transition and develop a distance measure. Feature-based methods generally consist two steps: feature extraction and classification \cite{fulcher2014highly}. As reviewed in \cite{Bagnall2017The}, bag-of-words based approaches, also called dictionary-based approaches, are an important branch of time series classification methods. There are many improved and modified versions of the original method, e.g., the bag of features (BOF) \cite{baydogan2013a}, bag of patterns (BOP) \cite{Jessica2012Rotation}, symbolic aggregate approximation-vector space model (SAXVSM) \cite{Senin2013SAX} and bag of SFA symbols (BOSS) \cite{Sch2015The}. Deep learning-based methods have also been studied in recent years. As reviewed in \cite{fawaz2019deep}, methods using fully convolutional network (FCN) \cite{wang2017time}, echo state network (ESN) \cite{aswolinskiy2018time}, long short-term memory recurrent neural network (LSTM) \cite{karim2018lstm} etc. have been developed. Considering the restricted number of training samples and drastic change of the time series with driver states, feature-based approach is exploited in this research. Similar to the idea of bag-of-words encoding, a driver can be characterized by a bag of driver states presented during car-following as we do in this study, where the main difference of the proposed approach is that the encoding and matching steps in common bag-of-words based methods are implicitly combined by directly using the joint posterior probability of online measurements based on the GMM based driver models.

\section{Approach}

\subsection{Background and Assumptions} \label{subsec_approach_background}

Driver identification can be formulated as a classification problem: given a sample $x \in \mathbf{R}^n$, predict its label $y \in Y$ where $Y$ is a label set with finite elements. In a car-following behavior based driver identification scenario, $x$ can be a feature vector extracted from a raw car-following sequence $S$ by a feature extractor $f$, i.e., $x = f(S)$, and $y$ is a driver's ID.

Generally speaking, the problem can be formulated using a generative model or a discriminative model. In the former way, $p(x|y)$ needs to be estimated, so that $y$ can be estimated by $p(y|x) = p(y)p(x|y)/\sum_y p(y)p(x|y)$, where $p(y)$ is a prior probability of labels which can be estimated using empirical probability and in some situations can be just assumed to be uniform, but $p(x|y)$ is usually in quite limited form (e.g., mixture of Gaussian) for tractability of computation, while in the discriminative way, $p(y|x)$ is directly modeled. In modern machine learning approaches, discriminative models are broadly used because very complicated and representative models (e.g., deep neural network) can be used to approximate $p(y|x)$ and learning process is straightforward by minimizing a loss function using gradient based numerical optimization algorithm.

However, for driver identification, a generative model is preferred, because in real world application, the label set is open, i.e., data of new driver will always be added, in which case a discriminative model need to retrain the model parameters using all driver's data, while for a generative model, we only need to estimate $p(x|y)$ for the new driver $y$ using his own data. To overcome the shortcoming of generative model, we can combine the idea of discriminative model by introducing learnable parameters in $f$, in which case $f$ can be denoted as $f(\,\cdot\,|\alpha)$. If there is a method to learn $\alpha$ in condition that $f(\,\cdot\,|\alpha)$ is differentiable with respect to $\alpha$, $f$ can be selected as representative as in discriminative models.

In our study, the framework of our approach is demonstrated in Fig.~\ref{framework}. The model is based on the following assumptions:
\begin{itemize}
\item There are $Q$ driver states shared by all drivers, characterizing the intra-driver heterogeneity.
\item There is a car-following feature space induced by $f$. Each car-following sequence $S$ can be projected to a feature vector $f(S)$ in the feature space. Each driver state corresponds to a Gaussian distribution in the feature space, which means the feature vector $f(S)$ generated by a driver under a certain driver state obeys the corresponding Gaussian distribution.
\item Each driver corresponds to a prior distribution over the driver states, called \emph{driver profile} in this study. Different driver profiles across drivers characterizing the inter-driver heterogeneity.
\end{itemize}

\subsection{Learning and Inference} \label{subsec_learn_and_infer}

As analyzed in \ref{subsec_approach_background}, three groups of parameters of the model need to be estimated from training data (car-following sequences with ground truth driver IDs):
\begin{itemize}
\item $\alpha$ in feature extractor $f(\,\cdot\,|\alpha)$.
\item Gaussian parameters of all driver states, i.e., $\mu_q$ and $\Sigma_q$, $q \in \{1, 2, \ldots, Q\}$.
\item Drive profile all drivers, i.e., $\omega_k = (\omega_{k,1}, \ldots, \omega_{k,Q})$, $k \in \{1, 2, \ldots, K\}$.
\end{itemize}

If all above parameters have been estimated in the learning phase, given a new observed sequence $S$, the posterior probability that it's generated by driver $k$ would be:
\begin{equation}
\label{posterior_prob}
P(k|S) = \frac{p_k(x)}{\sum_{k'=1}^K p_{k'}(x)}
\end{equation}
where
\begin{eqnarray}
x &=& f(S|\alpha) \label{def_x}\\
p_k(x) &=& \sum_{q=1}^Q\omega_{k,q}p_{\boldsymbol{N}}(x|\mu_q, \Sigma_q) \label{def_pk}
\end{eqnarray}
where $p_{\boldsymbol{N}}(\,\cdot\,| \mu, \Sigma)$ represents the probability density function of Gaussian distribution $\boldsymbol{N}(\mu, \Sigma)$, i.e.,
\begin{equation}
p_{\boldsymbol{N}}(x | \mu, \Sigma) = \frac{1}{(2\pi)^{n/2}\left|\Sigma\right|^{1/2}}e^{-\frac{1}{2}(x-\mu)^T\Sigma^{-1}(x-\mu)}
\end{equation}
Thus, the driver identity $k^*$ can be inferred as the ID with maximum posterior probability:
\begin{equation} \label{eqn_infer}
k^* = \argmax_k P(k|S)
\end{equation}
If multiple sequences of a certain driver are observed, by assuming they are independent of each other, the driver identity $k^*$ can be inferred as follows:
\begin{equation}  \label{eqn_multi_infer}
k^* = \argmax_k \prod_n P(k|S^{(n)})
\end{equation}

To learn the parameters, note that if $\alpha$ is given, following the methodology of generative model, we can simply estimate other parameters (driver states and driver profiles) by maximizing the likelihood of all $x = f(S|\alpha)$ in training set:
\begin{equation} \label{eqn_mu_Sigma_omega_star}
\left(\mu_{1:Q}^*, \Sigma_{1:Q}^*, \omega_{1:K}^*\right) = \argmax_{\mu_{1:Q},\Sigma_{1:Q},\omega_{1:K}}\prod_{k=1}^K\prod_{n=1}^{N_k}p_k(x_k^{(n)})
\end{equation}
where the subscript $k$ and superscript $(n)$ in $x_k^{(n)}$ represent the term corresponds to the $n$th sample of driver $k$ in training set. The usage will also appear in remaining part of the article and will not be explained if there's no ambiguity.

But how can $\alpha$ be jointly learned? Note that $\alpha$ controls the projection of car-following sequences $S$ to feature space, different $\alpha$ may lead to different driver states and driver profiles, thus different ability to distinguish different drivers. For the task of driver identification, we need to find out an $\alpha$, so that in the feature space induced by $f(\,\cdot\,|\alpha)$, the learned driver states and driver profiles can best help distinguish different drivers, i.e., the joint posterior probability of ground truth driver IDs should be maximized:
\begin{equation}
\label{learning_obj_vmax}
\alpha^* = \argmax_{\alpha} \prod_{k=1}^K\prod_{n=1}^{N_k} P^*(k|S^{(n)},\alpha)
\end{equation}
where
\begin{equation} \label{eqn_P_star}
P^*(k|S,\alpha) = P(k|S, \alpha, \mu_{1:Q}^*(\alpha), \Sigma_{1:Q}^*(\alpha), \omega_{1:K}^*(\alpha))
\end{equation}
The right hand of Eqn.~(\ref{eqn_P_star}) follows the same definition as Eqn.~(\ref{posterior_prob}), except that it appends all parameter dependencies for clarity. In Eqn.~(\ref{posterior_prob}), $P(k|S)$ is dependent on $p_k(x), k\in\{1,\ldots,K\}$; in Eqn.~(\ref{def_x}), $x$ is dependent on $\alpha$; in Eqn.~(\ref{def_pk}), $p_k(\,\cdot\,)$ is dependent on $\mu_{1:Q}, \Sigma_{1:Q}, \omega_{k}$. Thus, $P(k|S)$ could be written as $P(k|S, \alpha, \mu_{1:Q}, \Sigma_{1:Q}, \omega_{1:K})$. In addition, in Eqn.~(\ref{eqn_P_star}), we use the expression $\mu_{1:Q}^*(\alpha), \Sigma_{1:Q}^*(\alpha), \omega_{1:K}^*(\alpha)$ to indicate that they are optimal parameters obtained under a given $\alpha$, as described in the previous paragraph. Solving Eqn.~(\ref{learning_obj_vmax}), the parameters for inference are obtained as follows:
\begin{equation}
\alpha^*, \mu_{1:Q}^*(\alpha^*), \Sigma_{1:Q}^*(\alpha^*), \omega_{1:K}^*(\alpha^*)
\end{equation}

\subsection{Optimization Problem Solving} \label{subsec_optimization}

To facilitate the description of the optimization algorithm, we first define notations as follows:
\begin{equation}
\begin{split}
&L_k^{(n)}(\alpha_1, \alpha_2) = \\
&-\log P(k|S_k^{(n)}, \alpha_1, \mu_{1:Q}^*(\alpha_2), \Sigma_{1:Q}^*(\alpha_2), \omega_{1:K}^*(\alpha_2))
\end{split}
\end{equation}
\begin{equation}
L(\alpha_1, \alpha_2) = \sum_{k=1}^K\sum_{n=1}^{n_k}L_k^{(n)}(\alpha_1, \alpha_2)
\end{equation}
Note that following this notation, Eqn.~(\ref{learning_obj_vmax}) is equivalent to the following problem:
\begin{equation} \label{learning_obj_vmin}
\alpha^* = \argmin_{\alpha}L(\alpha, \alpha)
\end{equation}

In order to perform numerical optimization of $L(\alpha,\alpha)$, we need $L$'s gradient with respect to $\alpha$:
\begin{equation}
\frac{\partial L(\alpha, \alpha)}{\partial \alpha} = \left.\frac{\partial L(\alpha_1, \alpha)}{\partial \alpha_1}\right|_{\alpha_1=\alpha} + \left.\frac{\partial L(\alpha, \alpha_2)}{\partial \alpha_2}\right|_{\alpha_2=\alpha}
\end{equation}
However, $\partial L(\alpha_1, \alpha_2)/\partial \alpha_2$ cannot be represented in closed form. In this study, we propose to simply use $\left.\partial L(\alpha_1, \alpha)/\partial \alpha_1\right|_{\alpha_1=\alpha}$ to approximate $\partial L(\alpha, \alpha)/\partial \alpha$. In experimental section, it will be demonstrated how the practice works. The basic outline of the algorithm to solve the optimization problem defined by Eqn.~(\ref{learning_obj_vmin}) is shown in Algorithm~\ref{algorithm_outer_loop}, with a reference to Algorithm~\ref{algorithm_EM} for solving the optimization problem defined by Eqn.~(\ref{eqn_mu_Sigma_omega_star}).

\begin{algorithm}
\caption{Basic Outline of Solving Eqn.~(\ref{learning_obj_vmin})}
\label{algorithm_outer_loop}
\begin{algorithmic}
\STATE
\STATE Initialize $\alpha$.
\STATE
\FOR{$i=1:N_{Iter}$}
\STATE
\STATE Refer to Algorithm~\ref{algorithm_EM}: solve the problem defined by Eqn.~(\ref{eqn_mu_Sigma_omega_star}) to get $\mu_{1:Q}^*(\alpha),\Sigma_{1:Q}^*(\alpha),\omega_{1:K}^*(\alpha)$, so that $L(\,\cdot\,,\alpha)$ can be obtained.
\STATE
\STATE Update $\alpha$:
\begin{equation}
\alpha \gets \alpha - r \cdot \left.\frac{\partial L(\alpha_1, \alpha)}{\partial \alpha_1}\right|_{\alpha_1=\alpha} \nonumber
\end{equation}
\STATE
\ENDFOR
\end{algorithmic}
\end{algorithm}

\begin{algorithm}
\caption{EM Algorithm for Solving Eqn.~(\ref{eqn_mu_Sigma_omega_star})}
\label{algorithm_EM}
\begin{algorithmic}
\STATE
\STATE Initialize $\mu_{1:Q},\Sigma_{1:Q},\omega_{1:K}$.
\STATE
\FOR{$i=1:N_{Iter}$}
\STATE
\STATE E-Step:
\begin{equation}
\gamma_k^{(n)}(q) = \frac{\omega_{k,q}p_{\boldsymbol{N}}(\tilde{x}_k^{(n)}|\mu_q, \Sigma_q)}{\sum_{q'=1}^Q \omega_{k,q'}p_{\boldsymbol{N}}(\tilde{x}_k^{(n)}|\mu_{q'}, \Sigma_{q'})} \nonumber
\end{equation}
\STATE
\STATE M-Step:
\begin{eqnarray}
\omega_{k,q}&=&\frac{M_{k,q}}{N_k}, \ k=1:k, \ q=1:Q \nonumber\\
\Sigma_q&=&\frac{\sum_{k=1}^K\sum_{n=1}^{N_k}\gamma_k^{(n)}(q)(\tilde{x}_k^{(n)}\!-\!\mu_q)(\tilde{x}_k^{(n)}\!-\!\mu_q)^T}{M_q}, \nonumber\\
&&q=1:Q \nonumber\\
\mu_q&=&\frac{\sum_{k=1}^K\sum_{n=1}^{N_k}\gamma_k^{(n)}(q)\tilde{x}_k^{(n)}}{M_q}, \ q=1:Q \nonumber
\end{eqnarray}
where $M_{k,q}=\sum_{n=1}^{N_k}\gamma_k^{(n)}(q)$, $M_q = \sum_{k=1}^K M_{k,q}$.
\STATE
\ENDFOR
\end{algorithmic}
\end{algorithm}

\subsection{Feature Extractor}

Implementation of $f(\,\cdot\,|\alpha)$ is described in this part. We define input car-following sequences as time series of fixed length $T$. Each frame of the sequence is 4-dimensional, containing:
\begin{itemize}
\item ego's longitudinal velocity $v$;
\item ego's longitudinal acceleration $a$;
\item longitudinal distance of leading vehicle to ego $h$;
\item longitudinal relative velocity of leading vehicle to ego $\dot{h}$.
\end{itemize}
Thus, a car-following sequence $S\in\mathbf{R}^{T\times 4}$ can be represented as:
\begin{equation}
S=[v_{1:T},a_{1:T},h_{1:T},\dot{h}_{1:T}]
\end{equation}

Given such a car-following sequence $S$, we first extract a hand-crafted feature vector $\tilde x$ as introduced below. First, commonly used features like mean ego velocity, mean car following distance, mean ego acceleration will be directly calculated as features. Considering that a driver may have different preferences during accelerating and decelerating, mean positive acceleration and mean negative acceleration are extracted separately.

In many relevant researches, time to collision (TTC) has been taken as an important feature indicating a driver's control of ego vehicle. A typical TTC of frame $j$ is calculated as:
\begin{equation}
TTC_{j}=\frac{h_{j}}{\dot{h}_{j}}
\end{equation} Note that the negative TTC are meaningless. In order to eliminate the influence of puny value of relative speed, we choose the harmonic mean of positive TTC as one of the features.

In addition, reaction time (RT) is also introduced to represent the driver's reaction to the change of lead vehicle's velocity, which is calculated as:
\begin{equation}
RT=\arg_{\tau}\max(\rho_{xy}(\tau)),\ \tau_{min}<\tau<\tau_{max}
\end{equation}
where $x$ is the velocity sequence $v_{ego}=v_{1:T}$ of ego, $y$ is the velocity sequence $v_{lead}=v_{1:T}+h'_{1:T}$ of leading vehicle, $\rho_{xy}(\tau)$ is the cross-correlation calculation of sequence $x$ and $y$ under time shift $\tau$, $\tau_{min}$ and $\tau_{max}$ are the set boundaries of $\tau$.
 In order to distinguish the situation of different levels of cross-correlation, the value of maximum $\rho_{xy}(\tau)$ is also taken as one of the features of the input sequence.

 The detailed definition of features are shown below, in which $\mathds{1}$ is an indicator function.
 \begin{framed}
 	\noindent
1. Mean Ego Velocity:
\begin{equation}
f_{1}=\frac{1}{T}\sum_{j=1}^{T}v_{j}
\end{equation}
2. Mean Car-following Distance:
\begin{equation}
f_{2}=\frac{1}{T}\sum_{j=1}^{T}h_{j}
\end{equation}
3. Mean Ego Acceleration:
\begin{equation} \label{eqn_mean_acc}
f_{3}=\frac{1}{T}\sum_{j=1}^{T}a_{j}
\end{equation}
4. Mean Positive Acceleration:
\begin{equation}
f_{4}=\frac{1}{\sum_{j}\mathds{1}(a_{j}>0)}\sum_{j=1}^{T}a_{j}\cdot\mathds{1}(a_{j}>0)
\end{equation}
5. Mean Negative Acceleration:
\begin{equation}
f_{5}=\frac{1}{\sum_{j}\mathds{1}(a_{j}<0)}\sum_{j=1}^{T}a_{j}\cdot\mathds{1}(a_{j}<0)
\end{equation}
6. Harmonic Mean of TTC:
\begin{equation} \label{eqn_ttc}
f_{6}=\frac{\sum_{j}\mathds{1}(\dot{h}_{j}>0)}{\sum_{1}^{T}\frac{\mathds{1}(\dot{h}_{j}>0)}{TTC_{j}}}
\end{equation}
7. Reaction Time:
\begin{equation}
f_{7}=\arg_{\tau}\max(\rho_{xy}(\tau)),\ \tau_{min}<\tau<\tau_{max}
\end{equation}
8. Maximum Cross-correlation under Reaction Time:
\begin{equation}
f_{8}=\max(\rho_{xy}(\tau)),\ \tau_{min}<\tau<\tau_{max}
\end{equation}
\end{framed}

With the definition above, a hand-crafted feature vector of the given sequence $S$ can be obtained as:
\begin{equation} \label{eqn_tilde_x}
\tilde x=(f_{1}, f_{2},\cdots,f_{8})^T \in \mathbf{R}^8.
\end{equation}
Then, a linear projection is applied on $\tilde x$ to get the final output of $f(\,\cdot\,|\alpha)$:
\begin{equation} \label{eqn_x}
x = A\tilde x, A\in\mathbf{R}^{M\times 8}, x\in \mathbf{R}^M, M < 8
\end{equation}
In summary, as for implementation of $f(\,\cdot\,|\alpha)$, $\alpha$ is the matrix $A$, and $f$ first map $S$ to $\tilde x$ by a hand-crafted feature extractor, then project it to $x$ in a lower dimensional feature space using $A$, which should be learned as described in \ref{subsec_learn_and_infer} and \ref{subsec_optimization}.

\subsection{Implementation Details}

\subsubsection{Normalization of $\tilde x$} \label{subsubsec_norm_tilde_x}
$\tilde x$ should first be standardized before projected to low dimensional feature space, i.e., each dimension minus mean value, and divided by standard deviation, where the mean value and standard deviation are estimated using $\tilde x$ extracted from training set. The operation makes each dimension of $\tilde x$ equally treated, and the projected points $x$ not globally drift when $A$ changes during optimization.

\subsubsection{Normalization of $A$} \label{subsubsec_norm_A}
According to Algorithm~\ref{algorithm_outer_loop}, $A$ will be updated along the gradient (see ``Update $\alpha$''). To prevent all absolute values in $A$ becomes to large or small, each row of $A$ is normalized by dividing its Euclidean length after each update. Note that the only impact made by this operation is rescaling each dimension of the low dimensional feature space, which will not affect the discriminability of feature points $x$ in the space.

Together with normalization of $\tilde x$ as described in \ref{subsubsec_norm_tilde_x}, such a normalization strategy of $A=(a_{ij})$ can help in measuring the contribution of each dimension of $\tilde x$ after all parameters are learned, as presented below:
\begin{equation}
C(f_j) = \sum_{i=1}^M a_{ij}^2
\end{equation}

\subsubsection{Adaptive learning rate $r$}
Large learning rate $r$ may cause problem in convergence, while small $r$ will decrease the efficiency of the learning process. We propose to use an adaptive learning rate. After update of $\alpha$ in Algorithm~\ref{algorithm_outer_loop} in each iteration, the loss $L(\alpha, \alpha)$ (see Eqn.~\ref{learning_obj_vmin}) is compared with the loss before update. If the loss decreases after the update, $\alpha$ will be increased to $\gamma_1 \alpha$, where $\gamma_1 > 1$. Otherwise, $\alpha$ will be decreased to $\gamma_2 \alpha$ where $\gamma_2 < 1$. What's more, to ensure stability, there is an upper bound of $\alpha$, denoted as $\bar{\alpha}$. If $\alpha$ is increased to a value larger than $\bar{\alpha}$, then $\alpha$ is set to $\bar{\alpha}$. In our implementation, we set $\gamma_1=1.1$, $\gamma_2=0.5$ and $\bar{\alpha}=0.1$.

\subsubsection{Best result cache}
Since there is no guarantee that the loss will decrease in each iteration of Algorithm~\ref{algorithm_outer_loop}, a practical solution would be that cache the minimal loss as well as the corresponding parameters and update it when smaller loss is found during the training process.

\subsubsection{Initialization of Algorithm~\ref{algorithm_EM}}
Algorithm~\ref{algorithm_EM} is proposed to update $\mu_{1:Q},\Sigma_{1:Q},\omega_{1:K}$ in each iteration of Algorithm~\ref{algorithm_outer_loop}. The adopted EM algorithm requires initialization of the parameters. In our implementation, for the first time Algorithm~\ref{algorithm_EM} runs, parameters are randomly initialized. Later, the parameters are initialized using the value calculated in the previous iteration of Algorithm~\ref{algorithm_outer_loop}.

\subsubsection{Iteration times of Algorithm~\ref{algorithm_EM}}
In our practice, it is found that there's no need to wait for a perfect convergence in Algorithm~\ref{algorithm_EM}. To accelerate the training process, we set $N_{Iter}=10$. And when $\alpha$ is finally decided in Algorithm~\ref{algorithm_outer_loop}, Algorithm~\ref{algorithm_EM} is run again with a large $N_{Iter}$ to get the final result.

\section{Experiments}

\subsection{Dataset}

Experiments are conducted using the data collected through naturalistic driving of an instrumented vehicle by different drivers on the 4th Ring Road of Beijing. From the ego vehicle's CAN bus, the ego's motion states such as speed and acceleration are collected; by using four horizontal 2D lidars and a software of vehicle detection and tracking \cite{Zhao2017On}, trajectories of surrounding vehicles, especially that of the leading vehicle during car-following, are collected as well. With these collected trajectories and treating the instrumented vehicle as the follower, raw car-following sequences are extracted by detecting the data segments meeting the following criterions during the period: 1) the leading vehicle's id does not change; 2) the relative distance to the leading vehicle is within 40 meters; 3) the sequence is longer than 25 seconds. Note that the raw car-following sequences are of variable lengths, from 30 seconds to 180 seconds in our practice. A dataset containing car-following sequences of 8 drivers is prepared for the following experiments, which covers scenarios of various driving speed. Data of each driver is divided into training and testing set for evaluation of the proposed method. More detailed information about the dataset is listed in Table~\ref{tab_dataset}.

\begin{table}[h]
	\centering
	\caption[Table 1]{Dataset}\label{tab_dataset}
	
		\begin{tabular}{|c|c|c|c|c|c|c|}
			\hline
			ID    & \begin{tabular}[c]{@{}c@{}}Total \\ Seq.\\ Num.\end{tabular} & \begin{tabular}[c]{@{}c@{}}Total\\ Seq.\\  Len.(s)\end{tabular} & \begin{tabular}[c]{@{}c@{}}Train\\  Seq.\\  Num.\end{tabular} & \begin{tabular}[c]{@{}c@{}}Train \\ Seq.\\ Len.(s)\end{tabular} & \begin{tabular}[c]{@{}c@{}}Test \\ Seq.\\ Num.\end{tabular} & \begin{tabular}[c]{@{}c@{}}Test \\ Seq.\\ Len.(s)\end{tabular} \\ \hline
			D1    & 109                                                          & 5068.3                                                          & 87                                                            & 3996.2                                                          & 22                                                          & 1072.1                                                         \\ \hline
			D2    & 93                                                           & 4283.9                                                          & 74                                                            & 3305.2                                                          & 19                                                          & 978.7                                                          \\ \hline
			D3    & 109                                                          & 4828.9                                                          & 87                                                            & 3718.9                                                          & 22                                                          & 1110                                                           \\ \hline
			D4    & 101                                                          & 5241.9                                                          & 80                                                            & 4261.3                                                          & 21                                                          & 980.6                                                          \\ \hline
			D5    & 69                                                           & 4516.8                                                          & 55                                                            & 3663.2                                                          & 14                                                          & 853.6                                                          \\ \hline
			D6    & 87                                                           & 5451.3                                                          & 69                                                            & 4385.3                                                          & 18                                                          & 1066                                                           \\ \hline
			D7    & 105                                                          & 5009.5                                                          & 84                                                            & 3995.1                                                          & 21                                                          & 1014.4                                                         \\ \hline
			D8    & 95                                                           & 3985.6                                                          & 76                                                            & 3350.3                                                          & 16                                                          & 635.3                                                          \\ \hline
			Total & 768                                                          & 38386.2                                                         & 612                                                           & 30675.5                                                         & 153                                                         & 7710.7                                                         \\ \hline
		\end{tabular}
	
\end{table}

\subsection{3-Driver Experiment---Model and Algorithm Visualization}\label{subsec_3_driver_exp}

\begin{figure*}[]
  \begin{center}
    \includegraphics[keepaspectratio=true,width=0.8\linewidth]{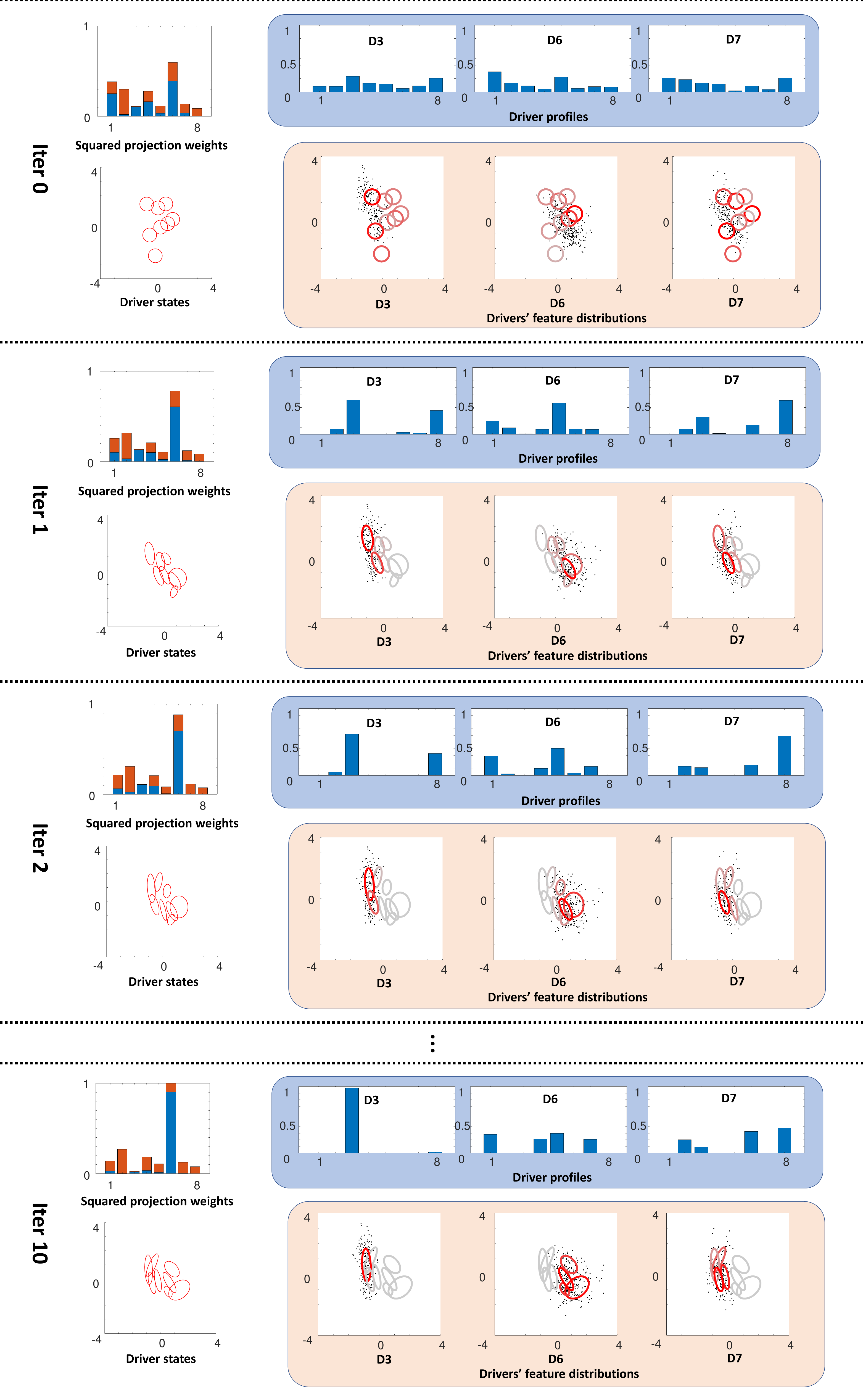}
  \end{center}
\caption{3-driver experiment: model parameters at different training iterations are visualized. Please refer to \ref{subsec_3_driver_exp} for detailed explanation of subfigures.}
\label{train_visualization}
\end{figure*}

\begin{figure}[tb]
  \begin{center}
    \includegraphics[keepaspectratio=true,width=1\linewidth]{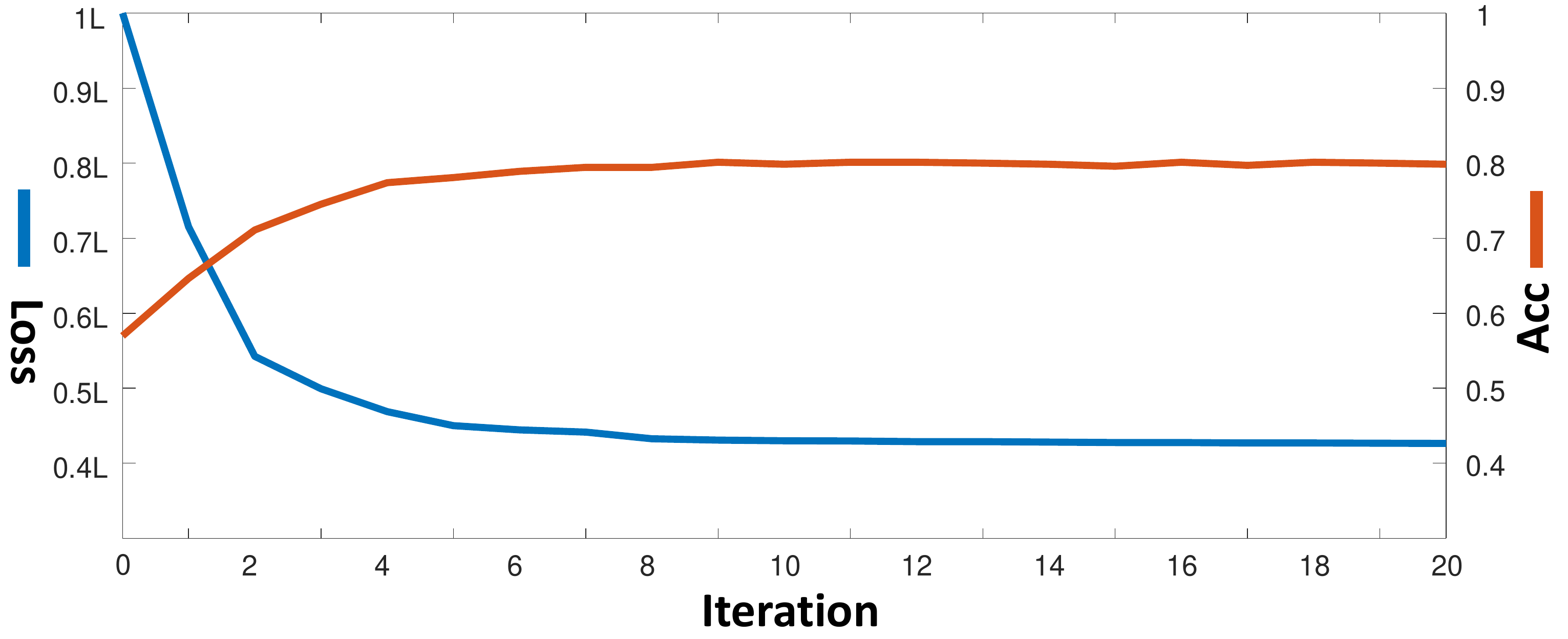}
  \end{center}
\caption{3-driver experiment: loss and accuracy evolution during training process.}
\label{Loss}
\end{figure}

In this part, some visualization results are presented to demonstrate how the proposed model and training algorithm works. For ease of visualization, the dimension of feature space is set to be $M=2$, and only training data of three drivers (D3, D6 and D7) are used. In addition, we set $Q=8$ and $T=15s$.

In Fig.~\ref{train_visualization}, model parameters in iteration 0 (initialization), iteration 1, iteration 2 and iteration 10 are visualized. For each iteration step, the subfigure \emph{Squared projection weights} shows $a_{ij}^2$ (denoting $A=(a_{ij})$ and please refer to \ref{subsubsec_norm_A} for the meaning of $a_{ij}^2$) using stacked histograms, where each $i$ corresponds to a specific color (blue for $i=1$ and orange for $i=2$) and different $j$s are distributed along x-axis; the subfigure \emph{Driver states} shows Gaussian parameters ($\mu_q, \Sigma_q, q=1,\cdots,8$) of all driver states using density contours (one ellipse for each Gaussian distribution); the subfigure \emph{Driver profiles} shows $\omega_k, k=1,2,3$ using 3 histograms, of which each corresponds to a driver; and the subfigure \emph{Drivers' feature distributions} shows the Gaussian mixture of each driver by combining Gaussian kernels in \emph{Driver states} and corresponding weights in \emph{Driver profiles} (refer to \ref{def_pk}) as well as the feature points ($x$) of each driver in training data, where higher saturation of the ellipse color represents higher weight (prior probability) of the driver on the corresponding driver states. Fig.~\ref{Loss} shows how the loss decreases and how the accuracy increases on training set as the iteration progresses.

From the result, we see that the loss decreases fast in the first several iteration steps and all model parameters evolves significantly to achieve the optimization goal of decreasing the loss. What's interesting is that at iteration 10, where the loss almost converges, in subfigure \emph{Squared projection weights}, the height of the 6th bar is obviously higher than others, while the 3rd bar is almost invisible. According to analysis in \ref{subsubsec_norm_A}, the height of each bar in subfigure \emph{Squared projection weights} actually measures the contribution of the corresponding dimension in $\tilde x$. According to Eqn.~(\ref{eqn_tilde_x}), (\ref{eqn_ttc}) and (\ref{eqn_mean_acc}), the $\tilde x$'s 6th dimension $f_6$ is harmonic mean of TTC, and the 3rd dimension is mean ego's acceleration, which indicates that the TTC is a very significant feature for discriminating the three drivers, while the mean ego's acceleration is helpless so that $A$ is optimized to discard almost all information of the dimension. By further observing the figure, we notice that the height of blue bar is almost $0$ except that at the 6th dimension, which means that $a_{1j}\approx 0 \text{ for } j\neq 6, a_{1j}\approx 1 \text{ for } j=6$. Thus, according to Eqn.~(\ref{eqn_x}), the first dimension of $x$ almost equals to $f_6$, which means that the x-axis of subfigure \emph{Drivers' feature distributions} actually represent $f_6$ of training samples. However, from the subfigure of iteration 10, we can find that only driver 6 can be separated well along x-axis from driver 3 and driver 7, while driver 3 and driver 7's feature points almost share the same distribution along x-axis, and they are actually separated along y-axis. Now we get an insight on the role that each manually extracted feature plays in this case: TTC helps to discriminate driver 6 from other drivers, while driver 3 and driver 7 are discriminated from each other by other features except mean ego's acceleration.

\begin{figure}[ht]
  \begin{center}
    \includegraphics[keepaspectratio=true,width=1\linewidth]{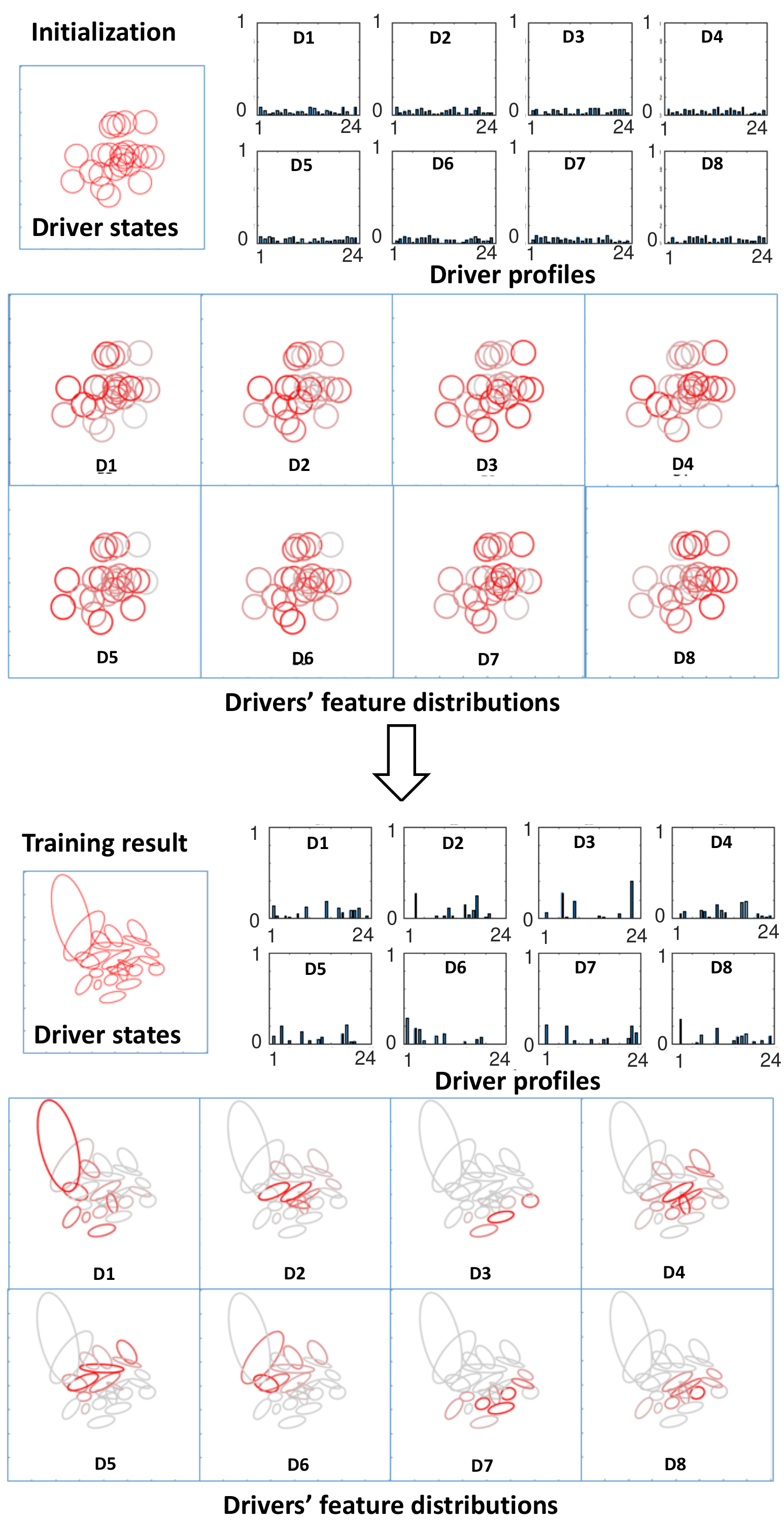}
  \end{center}
\caption{8-driver experiment: model parameters at initialization and after training. Hyper-parameters: $M=2$, $Q=24$, $T=15s$. Please refer to \ref{subsec_3_driver_exp} for detailed explanation of subfigures.}
\label{8exp}
\end{figure}

\subsection{8-Driver Experiment---Quantitative Evaluation} \label{subsec_quantitative_eval}

\subsubsection{Experimental design}

From the original data set, 80\% of the sequences of each driver's data are randomly selected for training, while the remaining 20\% sequences are for testing. Note that there are several hyperparameters of the method that may affect performance. First of all, the dimension of car-following behavior feature vector $M$ and the number of driver states $Q$ controls the capacity of the model. Thus, the following problem should be studied:
\begin{itemize}
	\item What is the suitable $M$ and $Q$ for driver identification?
\end{itemize}

Since the original data sequences are of variable lengths, resampling is conducted to convert them as sequences with uniform length $T$. As for resampling, the following two problems should be studied:
\begin{itemize}
	\item What is the suitable sequence length $T$ for driver identification?
	\item What is the suitable way for resampling, with overlap or not?
\end{itemize}

Experiments are first conducted on various combinations of $M$ and $Q$ with a fixed $T$ and non-overlapped resampling. Then, by fixing $M$ and $Q$, results generated with various $T$ and overlap rate are demonstrated and analysed. Finally, experiments of driver identification based on multiple car-following sequences are conducted to answer the following question:
\begin{itemize}
	\item How much improvement can be achieved by using multiple sequences for driver identification?
\end{itemize}
Since there are random factors that may affect the result (e.g., parameters initialization in learning process), each experiment was repeated 10 times with different random seed, and all results presented are average results over the repeated experiments.

\subsubsection{Feature space dimension and driver states number}

\begin{table}[tb]
	\centering
	\caption{8-driver experiment: average accuracy under various $(Q,M)$.}
	\begin{tabular}{|c|c|c|c|c|c|c|c|}
		\hline
		\multicolumn{8}{|c|}{Training Accuracy}         \\ \hline
		Q\verb|\|M&2&3&4&5&6&7&8\\ \hline
        8&\cellcolor[rgb]{1.00,0.72,0.72}0.394&\cellcolor[rgb]{1.00,0.75,0.75}0.388&\cellcolor[rgb]{1.00,0.63,0.63}0.408&\cellcolor[rgb]{1.00,0.78,0.78}0.382&\cellcolor[rgb]{1.00,0.75,0.75}0.386&\cellcolor[rgb]{1.00,0.91,0.91}0.347&\cellcolor[rgb]{1.00,0.97,0.97}0.325\\ \hline
        12&\cellcolor[rgb]{1.00,0.61,0.61}0.412&\cellcolor[rgb]{1.00,0.56,0.56}0.420&\cellcolor[rgb]{1.00,0.67,0.67}0.402&\cellcolor[rgb]{1.00,0.48,0.48}0.431&\cellcolor[rgb]{1.00,0.76,0.76}0.386&\cellcolor[rgb]{1.00,0.84,0.84}0.366&\cellcolor[rgb]{1.00,0.83,0.83}0.370\\ \hline
        16&\cellcolor[rgb]{1.00,0.68,0.68}0.400&\cellcolor[rgb]{1.00,0.53,0.53}0.423&\cellcolor[rgb]{1.00,0.51,0.51}0.427&\cellcolor[rgb]{1.00,0.49,0.49}0.429&\cellcolor[rgb]{1.00,0.53,0.53}0.423&\cellcolor[rgb]{1.00,0.70,0.70}0.397&\cellcolor[rgb]{1.00,0.71,0.71}0.395\\ \hline
        20&\cellcolor[rgb]{1.00,0.72,0.72}0.393&\cellcolor[rgb]{1.00,0.46,0.46}0.433&\cellcolor[rgb]{1.00,0.46,0.46}0.434&\cellcolor[rgb]{1.00,0.43,0.43}0.438&\cellcolor[rgb]{1.00,0.58,0.58}0.417&\cellcolor[rgb]{1.00,0.69,0.69}0.399&\cellcolor[rgb]{1.00,0.65,0.65}0.405\\ \hline
        24&\cellcolor[rgb]{1.00,0.65,0.65}0.406&\cellcolor[rgb]{1.00,0.33,0.33}0.450&\cellcolor[rgb]{1.00,0.51,0.51}0.427&\cellcolor[rgb]{1.00,0.40,0.40}0.442&\cellcolor[rgb]{1.00,0.42,0.42}0.438&\cellcolor[rgb]{1.00,0.83,0.83}0.368&\cellcolor[rgb]{1.00,0.64,0.64}0.407\\ \hline
        28&\cellcolor[rgb]{1.00,0.72,0.72}0.393&\cellcolor[rgb]{1.00,0.49,0.49}0.429&\cellcolor[rgb]{1.00,0.27,0.27}0.457&\cellcolor[rgb]{1.00,0.42,0.42}0.439&\cellcolor[rgb]{1.00,0.34,0.34}0.449&\cellcolor[rgb]{1.00,0.41,0.41}0.440&\cellcolor[rgb]{1.00,0.75,0.75}0.387\\ \hline
        32&\cellcolor[rgb]{1.00,0.55,0.55}0.421&\cellcolor[rgb]{1.00,0.42,0.42}0.439&\cellcolor[rgb]{1.00,0.02,0.02}0.484&\cellcolor[rgb]{1.00,0.31,0.31}0.452&\cellcolor[rgb]{1.00,0.26,0.26}0.458&\cellcolor[rgb]{1.00,0.33,0.33}0.451&\cellcolor[rgb]{1.00,0.49,0.49}0.429\\ \hline
        36&\cellcolor[rgb]{1.00,0.49,0.49}0.430&\cellcolor[rgb]{1.00,0.50,0.50}0.427&\cellcolor[rgb]{1.00,0.19,0.19}0.466&\cellcolor[rgb]{1.00,0.12,0.12}0.473&\cellcolor[rgb]{1.00,0.42,0.42}0.439&\cellcolor[rgb]{1.00,0.48,0.48}0.431&\cellcolor[rgb]{1.00,0.77,0.77}0.383\\ \hline
        40&\cellcolor[rgb]{1.00,0.65,0.65}0.406&\cellcolor[rgb]{1.00,0.36,0.36}0.447&\cellcolor[rgb]{1.00,0.14,0.14}0.471&\cellcolor[rgb]{1.00,0.00,0.00}0.486&\cellcolor[rgb]{1.00,0.01,0.01}0.485&\cellcolor[rgb]{1.00,0.10,0.10}0.476&\cellcolor[rgb]{1.00,0.57,0.57}0.418\\ \hline
        \hline
        \multicolumn{8}{|c|}{Testing Accuracy}         \\ \hline
		Q\verb|\|M&2&3&4&5&6&7&8\\ \hline
        8&\cellcolor[rgb]{1.00,0.59,0.59}0.415&\cellcolor[rgb]{1.00,0.70,0.70}0.397&\cellcolor[rgb]{1.00,0.75,0.75}0.388&\cellcolor[rgb]{1.00,0.75,0.75}0.387&\cellcolor[rgb]{1.00,0.87,0.87}0.358&\cellcolor[rgb]{1.00,0.97,0.97}0.320&\cellcolor[rgb]{1.00,0.98,0.98}0.317\\ \hline
        12&\cellcolor[rgb]{1.00,0.37,0.37}0.445&\cellcolor[rgb]{1.00,0.53,0.53}0.424&\cellcolor[rgb]{1.00,0.66,0.66}0.403&\cellcolor[rgb]{1.00,0.65,0.65}0.405&\cellcolor[rgb]{1.00,0.81,0.81}0.374&\cellcolor[rgb]{1.00,0.96,0.96}0.326&\cellcolor[rgb]{1.00,0.89,0.89}0.352\\ \hline
        16&\cellcolor[rgb]{1.00,0.70,0.70}0.396&\cellcolor[rgb]{1.00,0.63,0.63}0.409&\cellcolor[rgb]{1.00,0.67,0.67}0.402&\cellcolor[rgb]{1.00,0.79,0.79}0.379&\cellcolor[rgb]{1.00,0.80,0.80}0.376&\cellcolor[rgb]{1.00,0.94,0.94}0.338&\cellcolor[rgb]{1.00,0.93,0.93}0.339\\ \hline
        20&\cellcolor[rgb]{1.00,0.82,0.82}0.373&\cellcolor[rgb]{1.00,0.34,0.34}0.449&\cellcolor[rgb]{1.00,0.64,0.64}0.407&\cellcolor[rgb]{1.00,0.81,0.81}0.375&\cellcolor[rgb]{1.00,0.82,0.82}0.372&\cellcolor[rgb]{1.00,0.89,0.89}0.354&\cellcolor[rgb]{1.00,0.93,0.93}0.342\\ \hline
        24&\cellcolor[rgb]{1.00,0.78,0.78}0.381&\cellcolor[rgb]{1.00,0.35,0.35}0.448&\cellcolor[rgb]{1.00,0.72,0.72}0.393&\cellcolor[rgb]{1.00,0.69,0.69}0.397&\cellcolor[rgb]{1.00,0.86,0.86}0.361&\cellcolor[rgb]{1.00,0.99,0.99}0.308&\cellcolor[rgb]{1.00,0.90,0.90}0.351\\ \hline
        28&\cellcolor[rgb]{1.00,0.78,0.78}0.380&\cellcolor[rgb]{1.00,0.81,0.81}0.375&\cellcolor[rgb]{1.00,0.54,0.54}0.422&\cellcolor[rgb]{1.00,0.78,0.78}0.381&\cellcolor[rgb]{1.00,0.80,0.80}0.376&\cellcolor[rgb]{1.00,0.94,0.94}0.338&\cellcolor[rgb]{1.00,1.00,1.00}0.288\\ \hline
        32&\cellcolor[rgb]{1.00,0.70,0.70}0.397&\cellcolor[rgb]{1.00,0.72,0.72}0.392&\cellcolor[rgb]{1.00,0.62,0.62}0.411&\cellcolor[rgb]{1.00,0.87,0.87}0.360&\cellcolor[rgb]{1.00,0.74,0.74}0.388&\cellcolor[rgb]{1.00,0.68,0.68}0.401&\cellcolor[rgb]{1.00,0.86,0.86}0.362\\ \hline
        36&\cellcolor[rgb]{1.00,0.71,0.71}0.395&\cellcolor[rgb]{1.00,0.79,0.79}0.379&\cellcolor[rgb]{1.00,0.86,0.86}0.362&\cellcolor[rgb]{1.00,0.75,0.75}0.388&\cellcolor[rgb]{1.00,0.89,0.89}0.353&\cellcolor[rgb]{1.00,0.88,0.88}0.357&\cellcolor[rgb]{1.00,1.00,1.00}0.298\\ \hline
        40&\cellcolor[rgb]{1.00,0.91,0.91}0.349&\cellcolor[rgb]{1.00,0.75,0.75}0.387&\cellcolor[rgb]{1.00,0.76,0.76}0.385&\cellcolor[rgb]{1.00,0.88,0.88}0.358&\cellcolor[rgb]{1.00,0.68,0.68}0.400&\cellcolor[rgb]{1.00,0.88,0.88}0.358&\cellcolor[rgb]{1.00,0.94,0.94}0.335\\ \hline
	\end{tabular}
    \label{QM_results}
\end{table}

Tab.~\ref{QM_results} shows the training and testing results with various combinations of feature space dimension $M$ and driver state number $Q$. Sequences are resampled to length $T=15s$ without overlap. For a intuitive understanding of the effectiveness of the training process on the 8-driver dataset, Fig.~\ref{8exp} shows the model parameters at initialization and after training with $M=2$ and $Q=24$, where we see that each driver has a distinctive driver profile and feature distribution after training.

From Tab.~\ref{QM_results}, we can find that there is roughly an accuracy peak region in $(M,Q)$ space: $M\in[4,6]$, $Q\in[24,40]$, where the average accuracy is higher than surrounding $(M,Q)$ combinations, and accuracy tends to decrease when $(M,Q)$ go far from the region. Similarly, in testing results, there is also such a region: roughly the upper triangle part of $M\in[2,4], Q\in[12,28]$.

Since larger $M$ and $Q$ will increase the capacity of the model and lead to overfitting to the training set, for a given training set, there should be a suitable $(M,Q)$ that generalize well, i.e., achieves good accuracy on testing set, and the testing accuracy peak region in $(M,Q)$ space indicates where the best $(M,Q)$ should lie in. We choose $M=4$ and $Q=28$ for the following experiments.

It is worth noting that for both training and testing results, when $M=8$, the accuracy is low for all enumerated $Q$. The fact indicates the benefit of introducing $A$ to project manually extracted feature vector to lower dimensional feature space.

\subsubsection{Car-following sequence length} \label{sssec_seq_length}

\begin{table}[tb]
	\centering
	\caption{8-driver experiment: average accuracy under various $T$.}
	\begin{tabular}{|c|c|c|c|c|c|}
		\hline
        T&10s&15s&20s&25s&30s\\ \hline
        Training Accuracy&\cellcolor[rgb]{1.00,0.97,0.97}0.303&\cellcolor[rgb]{1.00,0.03,0.03}0.457&\cellcolor[rgb]{1.00,0.57,0.57}0.394&\cellcolor[rgb]{1.00,0.05,0.05}0.456&\cellcolor[rgb]{1.00,0.62,0.62}0.386\\ \hline
        Testing Accuracy&\cellcolor[rgb]{1.00,1.00,1.00}0.274&\cellcolor[rgb]{1.00,0.36,0.36}0.422&\cellcolor[rgb]{1.00,0.85,0.85}0.342&\cellcolor[rgb]{1.00,0.87,0.87}0.339&\cellcolor[rgb]{1.00,1.00,1.00}0.277\\ \hline
	\end{tabular}
    \label{seq_length_results}
\end{table}

Resampling length is an important factor that could have an influence on performance. If the resampling length is too short, some features cannot be extracted correctly, e.g., reaction time, which actually require some change of speed in the sequence and if the speed is almost constant, the best correlation point found will not be reliable. On the other hand, if the resampling length is too long, different driver states might be mixed in a sequence which makes the extracted feature no longer a behavior descriptor under a certain driver state.

Tab.~\ref{seq_length_results} shows the average training and testing accuracy produced under various resampling length $T$, while other parameters stay fixed as $M=4$ and $Q=28$. From the results, we can see for both training and testing, $T=15s$ achieves the highest accuracy. Besides, testing accuracy drastically decrease as $T$ getting further from $15s$, which indicates that the model cannot generalize well with a improper resampling length.

\subsubsection{W/O overlap in sequence resampling}

\begin{table}[tb]
	\centering
	\caption{8-driver experiment: average accuracy under various sampling overlap ratio.}
	\begin{tabular}{|c|c|c|c|}
		\hline
		Overlap&Training&Training&Testing\\
        Ratio&Seq. Num.&Accuracy&Accuracy\\ \hline
        0.00&1856&\cellcolor[rgb]{1.00,0.67,0.67}0.457&\cellcolor[rgb]{1.00,0.95,0.95}0.422\\ \hline
        0.25&2451&\cellcolor[rgb]{1.00,0.55,0.55}0.467&\cellcolor[rgb]{1.00,0.72,0.72}0.453\\ \hline
        0.50&3543&\cellcolor[rgb]{1.00,0.31,0.31}0.483&\cellcolor[rgb]{1.00,0.73,0.73}0.452\\ \hline
	\end{tabular}
    \label{overlap_result}
\end{table}

In our approach, a long car-following sequence should be first resampled into several short car-following sequences of length $T$. Assuming the start points of sampled sequences are uniformly sampled, the temporal range of a sampled sequence should be:
\begin{equation}
[nT',nT'+T], T'\leq T, n=0,1,\ldots
\end{equation}
where $T'$ is the sampling interval between two sequential start points. Following this notation, an overlap ratio is defined as
\begin{equation}
r = \frac{T-T'}{T}
\end{equation}

Tab.~\ref{overlap_result} shows the result with different overlap ratios for training set, where hyper-parameters are fixed as $M=4$, $Q=28$, $T=15s$. It can be found out that sampling with overlap will bring slight improvement on training and testing results, but overlap with ratio 0.5 performs no better than overlap with ratio 0.25 on testing set. The result is within expectation. 1) Sampling with overlap will improve performance because higher overlap ratio will make more training samples and it can be regarded as a method of data augmentation for the problem. 2) As overlap ratio increases, the benefit will decrease, because the augmented data will be more similar with others when overlap ratio becomes higher (note that sampled starting points will be closer) and the increased overlap ratio cannot provide much novel information.

\subsubsection{Inference with multiple sequences}

\begin{table}[tb]
	\centering
	\caption{8-driver experiment: average testing accuracy with multiple sequences.}
	\begin{tabular}{|c|c|c|c|c|c|}
		\hline
        T&10s&15s&20s&25s&30s\\ \hline
        3-Seq Accuracy&\cellcolor[rgb]{1.00,0.98,0.98}0.382&\cellcolor[rgb]{1.00,0.75,0.75}0.575&\cellcolor[rgb]{1.00,0.86,0.86}0.506&\cellcolor[rgb]{1.00,0.97,0.97}0.390&\cellcolor[rgb]{1.00,0.99,0.99}0.361\\ \hline
        5-Seq Accuracy&\cellcolor[rgb]{1.00,0.92,0.92}0.453&\cellcolor[rgb]{1.00,0.61,0.61}0.644&\cellcolor[rgb]{1.00,0.75,0.75}0.576&\cellcolor[rgb]{1.00,0.95,0.95}0.418&\cellcolor[rgb]{1.00,0.97,0.97}0.390\\ \hline
        10-Seq Accuracy&\cellcolor[rgb]{1.00,0.71,0.71}0.597&\cellcolor[rgb]{1.00,0.09,0.09}0.823&\cellcolor[rgb]{1.00,0.46,0.46}0.706&\cellcolor[rgb]{1.00,0.81,0.81}0.541&\cellcolor[rgb]{1.00,0.83,0.83}0.526\\ \hline
	\end{tabular}
    \label{multi_seq_results}
\end{table}

In real world applications, a driver's car-following sequences can be collected in an online manner, and multiple sequences can be used to boost the inference performance of the model using Eqn.~(\ref{eqn_multi_infer}). We test the multi-sequence inference performance based on model and data in \ref{sssec_seq_length}, and inference accuracy using different number of testing sequences with various $T$ is presented in Tab.~\ref{multi_seq_results}, which can be regarded as an extended part of Tab.~\ref{seq_length_results}.

From the table, we see that for all sequence length $T$, inference accuracy with multiple sequences is higher than with single sequence (refer to Tab.~\ref{seq_length_results}), and the accuracy will increase as the number of sequences used for inference increases. For $T=15s$, the 10-sequence inference can achieve accuracy of $82.3\%$, which means for the 8 driver relevant to this study, if we observe a driver's car-following behavior for over $15s\times10=150s$, his/her identity can be recognized with an accuracy around $82.3\%$.

\subsection{Case Study---New Driver Registration}
\begin{table}[tb]
	\centering
	\caption{Case study: confusion matrix of case study models on testing data.}
	\begin{tabular}{|c|c|c|c|c|c|c|c|c|c|}
		\hline
		\multicolumn{10}{|c|}{A1 --- 7 drivers. Accuracy: 0.401.}         \\ \hline
		GT\verb|\|PR&D1&D2&D3&D4&D5&D6&D7&D8&recall\\ \hline
        D1&\cellcolor[rgb]{0.32,0.83,0.32}28&\cellcolor[rgb]{0.90,0.98,0.90}4&n/a&\cellcolor[rgb]{0.90,0.98,0.90}4&\cellcolor[rgb]{0.93,0.98,0.93}3&\cellcolor[rgb]{0.85,0.96,0.85}6&\cellcolor[rgb]{0.76,0.94,0.76}10&\cellcolor[rgb]{0.85,0.96,0.85}6&\cellcolor[rgb]{1.00,0.64,0.64}0.459\\ \hline
        D2&\cellcolor[rgb]{0.88,0.97,0.88}5&\cellcolor[rgb]{0.71,0.93,0.71}12&n/a&\cellcolor[rgb]{0.71,0.93,0.71}12&\cellcolor[rgb]{0.83,0.96,0.83}7&\cellcolor[rgb]{0.90,0.98,0.90}4&\cellcolor[rgb]{0.95,0.99,0.95}2&\cellcolor[rgb]{0.88,0.97,0.88}5&\cellcolor[rgb]{1.00,0.89,0.89}0.255\\ \hline
        D3&n/a&n/a&n/a&n/a&n/a&n/a&n/a&n/a&n/a\\ \hline
        D4&\cellcolor[rgb]{0.90,0.98,0.90}4&\cellcolor[rgb]{0.71,0.93,0.71}12&n/a&\cellcolor[rgb]{0.59,0.90,0.59}17&\cellcolor[rgb]{0.93,0.98,0.93}3&\cellcolor[rgb]{0.88,0.97,0.88}5&\cellcolor[rgb]{0.76,0.94,0.76}10&\cellcolor[rgb]{0.68,0.92,0.68}13&\cellcolor[rgb]{1.00,0.88,0.88}0.266\\ \hline
        D5&\cellcolor[rgb]{0.95,0.99,0.95}2&\cellcolor[rgb]{0.78,0.95,0.78}9&n/a&\cellcolor[rgb]{0.93,0.98,0.93}3&\cellcolor[rgb]{0.29,0.82,0.29}29&\cellcolor[rgb]{0.61,0.90,0.61}16&\cellcolor[rgb]{0.93,0.98,0.93}3&\cellcolor[rgb]{0.95,0.99,0.95}2&\cellcolor[rgb]{1.00,0.65,0.65}0.453\\ \hline
        D6&\cellcolor[rgb]{0.85,0.96,0.85}6&\cellcolor[rgb]{0.73,0.93,0.73}11&n/a&\cellcolor[rgb]{0.85,0.96,0.85}6&\cellcolor[rgb]{0.59,0.90,0.59}17&\cellcolor[rgb]{0.20,0.80,0.20}33&\cellcolor[rgb]{0.98,0.99,0.98}1&\cellcolor[rgb]{0.90,0.98,0.90}4&\cellcolor[rgb]{1.00,0.69,0.69}0.423\\ \hline
        D7&\cellcolor[rgb]{0.85,0.96,0.85}6&\cellcolor[rgb]{0.93,0.98,0.93}3&n/a&\cellcolor[rgb]{0.78,0.95,0.78}9&\cellcolor[rgb]{0.95,0.99,0.95}2&\cellcolor[rgb]{1.00,1.00,1.00}0&\cellcolor[rgb]{0.44,0.86,0.44}23&\cellcolor[rgb]{0.73,0.93,0.73}11&\cellcolor[rgb]{1.00,0.69,0.69}0.426\\ \hline
        D8&\cellcolor[rgb]{0.95,0.99,0.95}2&\cellcolor[rgb]{0.88,0.97,0.88}5&n/a&\cellcolor[rgb]{0.88,0.97,0.88}5&\cellcolor[rgb]{0.90,0.98,0.90}4&\cellcolor[rgb]{0.98,0.99,0.98}1&\cellcolor[rgb]{0.85,0.96,0.85}6&\cellcolor[rgb]{0.39,0.85,0.39}25&\cellcolor[rgb]{1.00,0.54,0.54}0.521\\ \hline
        \hline
        \multicolumn{10}{|c|}{A2 --- 7 drivers with D3 registered. Accuracy: 0.377.}         \\
        \multicolumn{10}{|c|}{Accuracy with D3: 0.377. Accuracy without D3: 0.365.}         \\ \hline
		GT\verb|\|PR&D1&D2&D3&D4&D5&D6&D7&D8&recall\\ \hline
        D1&\cellcolor[rgb]{0.39,0.85,0.39}25&\cellcolor[rgb]{0.90,0.98,0.90}4&\cellcolor[rgb]{0.76,0.94,0.76}10&\cellcolor[rgb]{0.90,0.98,0.90}4&\cellcolor[rgb]{0.93,0.98,0.93}3&\cellcolor[rgb]{0.85,0.96,0.85}6&\cellcolor[rgb]{0.88,0.97,0.88}5&\cellcolor[rgb]{0.90,0.98,0.90}4&\cellcolor[rgb]{1.00,0.71,0.71}0.410\\ \hline
        D2&\cellcolor[rgb]{0.88,0.97,0.88}5&\cellcolor[rgb]{0.71,0.93,0.71}12&\cellcolor[rgb]{0.98,0.99,0.98}1&\cellcolor[rgb]{0.71,0.93,0.71}12&\cellcolor[rgb]{0.83,0.96,0.83}7&\cellcolor[rgb]{0.90,0.98,0.90}4&\cellcolor[rgb]{0.98,0.99,0.98}1&\cellcolor[rgb]{0.88,0.97,0.88}5&\cellcolor[rgb]{1.00,0.89,0.89}0.255\\ \hline
        D3&\cellcolor[rgb]{0.83,0.96,0.83}7&\cellcolor[rgb]{1.00,1.00,1.00}0&\cellcolor[rgb]{0.41,0.85,0.41}24&\cellcolor[rgb]{1.00,1.00,1.00}0&\cellcolor[rgb]{0.98,0.99,0.98}1&\cellcolor[rgb]{1.00,1.00,1.00}0&\cellcolor[rgb]{0.68,0.92,0.68}13&\cellcolor[rgb]{0.85,0.96,0.85}6&\cellcolor[rgb]{1.00,0.62,0.62}0.471\\ \hline
        D4&\cellcolor[rgb]{0.90,0.98,0.90}4&\cellcolor[rgb]{0.71,0.93,0.71}12&\cellcolor[rgb]{0.78,0.95,0.78}9&\cellcolor[rgb]{0.59,0.90,0.59}17&\cellcolor[rgb]{0.93,0.98,0.93}3&\cellcolor[rgb]{0.88,0.97,0.88}5&\cellcolor[rgb]{0.98,0.99,0.98}1&\cellcolor[rgb]{0.68,0.92,0.68}13&\cellcolor[rgb]{1.00,0.88,0.88}0.266\\ \hline
        D5&\cellcolor[rgb]{0.95,0.99,0.95}2&\cellcolor[rgb]{0.80,0.95,0.80}8&\cellcolor[rgb]{0.93,0.98,0.93}3&\cellcolor[rgb]{0.93,0.98,0.93}3&\cellcolor[rgb]{0.29,0.82,0.29}29&\cellcolor[rgb]{0.61,0.90,0.61}16&\cellcolor[rgb]{0.98,0.99,0.98}1&\cellcolor[rgb]{0.95,0.99,0.95}2&\cellcolor[rgb]{1.00,0.65,0.65}0.453\\ \hline
        D6&\cellcolor[rgb]{0.85,0.96,0.85}6&\cellcolor[rgb]{0.73,0.93,0.73}11&\cellcolor[rgb]{0.98,0.99,0.98}1&\cellcolor[rgb]{0.85,0.96,0.85}6&\cellcolor[rgb]{0.59,0.90,0.59}17&\cellcolor[rgb]{0.20,0.80,0.20}33&\cellcolor[rgb]{1.00,1.00,1.00}0&\cellcolor[rgb]{0.90,0.98,0.90}4&\cellcolor[rgb]{1.00,0.69,0.69}0.423\\ \hline
        D7&\cellcolor[rgb]{0.93,0.98,0.93}3&\cellcolor[rgb]{0.93,0.98,0.93}3&\cellcolor[rgb]{0.63,0.91,0.63}15&\cellcolor[rgb]{0.78,0.95,0.78}9&\cellcolor[rgb]{0.95,0.99,0.95}2&\cellcolor[rgb]{1.00,1.00,1.00}0&\cellcolor[rgb]{0.71,0.93,0.71}12&\cellcolor[rgb]{0.76,0.94,0.76}10&\cellcolor[rgb]{1.00,0.92,0.92}0.222\\ \hline
        D8&\cellcolor[rgb]{0.95,0.99,0.95}2&\cellcolor[rgb]{0.88,0.97,0.88}5&\cellcolor[rgb]{0.90,0.98,0.90}4&\cellcolor[rgb]{0.88,0.97,0.88}5&\cellcolor[rgb]{0.90,0.98,0.90}4&\cellcolor[rgb]{0.98,0.99,0.98}1&\cellcolor[rgb]{0.93,0.98,0.93}3&\cellcolor[rgb]{0.41,0.85,0.41}24&\cellcolor[rgb]{1.00,0.57,0.57}0.500\\ \hline
        \hline
        \multicolumn{10}{|c|}{A3 --- 8 drivers. Accuracy: 0.426.}         \\ \hline
		GT\verb|\|PR&D1&D2&D3&D4&D5&D6&D7&D8&recall\\ \hline
        D1&\cellcolor[rgb]{0.29,0.82,0.29}29&\cellcolor[rgb]{0.85,0.96,0.85}6&\cellcolor[rgb]{0.88,0.97,0.88}5&\cellcolor[rgb]{0.93,0.98,0.93}3&\cellcolor[rgb]{0.85,0.96,0.85}6&\cellcolor[rgb]{0.88,0.97,0.88}5&\cellcolor[rgb]{0.88,0.97,0.88}5&\cellcolor[rgb]{0.95,0.99,0.95}2&\cellcolor[rgb]{1.00,0.61,0.61}0.475\\ \hline
        D2&\cellcolor[rgb]{0.88,0.97,0.88}5&\cellcolor[rgb]{0.90,0.98,0.90}4&\cellcolor[rgb]{1.00,1.00,1.00}0&\cellcolor[rgb]{0.80,0.95,0.80}8&\cellcolor[rgb]{0.51,0.88,0.51}20&\cellcolor[rgb]{0.98,0.99,0.98}1&\cellcolor[rgb]{0.90,0.98,0.90}4&\cellcolor[rgb]{0.88,0.97,0.88}5&\cellcolor[rgb]{1.00,0.99,0.99}0.085\\ \hline
        D3&\cellcolor[rgb]{0.93,0.98,0.93}3&\cellcolor[rgb]{1.00,1.00,1.00}0&\cellcolor[rgb]{0.05,0.76,0.05}39&\cellcolor[rgb]{1.00,1.00,1.00}0&\cellcolor[rgb]{0.98,0.99,0.98}1&\cellcolor[rgb]{1.00,1.00,1.00}0&\cellcolor[rgb]{0.83,0.96,0.83}7&\cellcolor[rgb]{0.98,0.99,0.98}1&\cellcolor[rgb]{1.00,0.00,0.00}0.765\\ \hline
        D4&\cellcolor[rgb]{0.90,0.98,0.90}4&\cellcolor[rgb]{0.66,0.91,0.66}14&\cellcolor[rgb]{0.88,0.97,0.88}5&\cellcolor[rgb]{0.68,0.92,0.68}13&\cellcolor[rgb]{0.85,0.96,0.85}6&\cellcolor[rgb]{0.85,0.96,0.85}6&\cellcolor[rgb]{0.90,0.98,0.90}4&\cellcolor[rgb]{0.71,0.93,0.71}12&\cellcolor[rgb]{1.00,0.93,0.93}0.203\\ \hline
        D5&\cellcolor[rgb]{0.85,0.96,0.85}6&\cellcolor[rgb]{0.88,0.97,0.88}5&\cellcolor[rgb]{1.00,1.00,1.00}0&\cellcolor[rgb]{0.98,0.99,0.98}1&\cellcolor[rgb]{0.00,0.75,0.00}41&\cellcolor[rgb]{0.80,0.95,0.80}8&\cellcolor[rgb]{1.00,1.00,1.00}0&\cellcolor[rgb]{0.93,0.98,0.93}3&\cellcolor[rgb]{1.00,0.30,0.30}0.641\\ \hline
        D6&\cellcolor[rgb]{0.80,0.95,0.80}8&\cellcolor[rgb]{0.88,0.97,0.88}5&\cellcolor[rgb]{1.00,1.00,1.00}0&\cellcolor[rgb]{0.93,0.98,0.93}3&\cellcolor[rgb]{0.39,0.85,0.39}25&\cellcolor[rgb]{0.22,0.80,0.22}32&\cellcolor[rgb]{0.98,0.99,0.98}1&\cellcolor[rgb]{0.90,0.98,0.90}4&\cellcolor[rgb]{1.00,0.71,0.71}0.410\\ \hline
        D7&\cellcolor[rgb]{0.88,0.97,0.88}5&\cellcolor[rgb]{0.95,0.99,0.95}2&\cellcolor[rgb]{0.63,0.91,0.63}15&\cellcolor[rgb]{0.93,0.98,0.93}3&\cellcolor[rgb]{0.93,0.98,0.93}3&\cellcolor[rgb]{1.00,1.00,1.00}0&\cellcolor[rgb]{0.49,0.87,0.49}21&\cellcolor[rgb]{0.88,0.97,0.88}5&\cellcolor[rgb]{1.00,0.74,0.74}0.389\\ \hline
        D8&\cellcolor[rgb]{0.95,0.99,0.95}2&\cellcolor[rgb]{0.93,0.98,0.93}3&\cellcolor[rgb]{0.93,0.98,0.93}3&\cellcolor[rgb]{0.78,0.95,0.78}9&\cellcolor[rgb]{0.90,0.98,0.90}4&\cellcolor[rgb]{1.00,1.00,1.00}0&\cellcolor[rgb]{0.83,0.96,0.83}7&\cellcolor[rgb]{0.51,0.88,0.51}20&\cellcolor[rgb]{1.00,0.70,0.70}0.417\\ \hline
	\end{tabular}
    \label{case_study_results}
\end{table}

As pointed out in \ref{subsec_approach_background}, the proposed approach partially follows the methodology of generative models, so that in order to incorporate a new driver in the model (new driver registration), the training process for updating model will only depend on the data of the new driver. Assuming that there is a car following dataset with ground truth driver Ids (as we used in the experiment but may be of much greater scale), from which all discriminative features and driver states can be learned, when there is a new driver to be registered, we just need to estimate his/her driver profile as defined in \ref{subsec_learn_and_infer}. According to Algorithm~\ref{algorithm_EM}, the estimation of each driver's profile is independent from data of other drivers, which means a new driver can be registered with only his/her own data.

An experiment is conducted to practice the proposed new driver registration method and examine its performance on the 8 driver dataset as used in \ref{subsec_quantitative_eval}. First, a model (A1) of 7 drivers is trained based on the training set of all drivers except driver 3 with $M=4$ and $Q=28$. Then, a model (A2) of 8 drivers is obtained by registering driver 3 to A1. Finally, a model (A3) of 8 drivers is directly trained based on the training set of all drivers. Confusion matrixes of the three models on testing set are shown in Tab.~\ref{case_study_results}. By comparing the results of A1 and A2, we see that after the registration of driver 3, there is a significant decrease of accuracy on the identification of original 7 drivers. It can be noticed that the degradation is mainly caused by the confusion between driver 3 and driver 7, because in A1, the recall of driver 7 is 0.426 while in A2, it decreases to 0.222 and 27.8\% of the samples are misclassified as driver 3. However, from the results of A3, we see that if all drivers are jointly trained, the same level of accuracy can be achieved, which means that the model is potentially capable of incorporating 8 drivers, but in this case, training with 7 drivers cannot get an optimal feature space and driver states representation for discriminating driver 3 from others.

The experiment shows that the proposed model is potentially able to register a new driver in an efficient way, but the condition under which a new registered driver will not degrade the performance much should be further studied.

\section{Conclusion and Future Work}

In this study, a model considering both intra- and inter-driver heterogeneity in car-following behavior is proposed as an approach of driver profiling and identification. It is assumed that all drivers share a pool of driver states; under each state a car-following data sequence obeys a specific Gaussian distribution in feature space; each driver has his/her own probability distribution over the states, called driver profile, which characterize the intra-driver heterogeneity, while the difference between the driver profile of different drivers depict the inter-driver heterogeneity. Thus, the driver profile can be used to distinguish a driver from others. Based on the assumption, a stochastic car-following model is proposed to take both intra-driver and inter-driver heterogeneity into consideration, and a method is proposed to jointly learn parameters in behavioral feature extractor, driver states and driver profiles.

Experiments demonstrate the performance of the proposed method in driver identification on naturalistic car-following data. A 3-driver experiment is carried out to visualize the model evolution in training process, and demonstrate how to gain a insight of driver behavior difference by analyzing the learned parameters; experiments on 8-driver dataset are conducted for quantitative analysis on how hyper-parameters impact the performance of the approach, and accuracy of 82.3\% is achieved by using 10 car-following sequences of duration 15 seconds for online inference; a case study is carried out to demonstrate the potential to fast register a new driver to the existing model, and the performance should be further improved.

In the future, experiments on dataset of larger scale will be carried out to test the performance of the approach. With more data, feature extractor can be defined using much more complex and expressive models such as LSTM or CNN, and the performance of new driver registration
method is expected to be boosted and should be further analyzed.

\ifCLASSOPTIONcaptionsoff
  \newpage
\fi

\bibliographystyle{IEEEtran}
\bibliography{IEEEabrv,ref}

\begin{IEEEbiography}[{\includegraphics[width=1in,height=1.25in,clip,keepaspectratio]{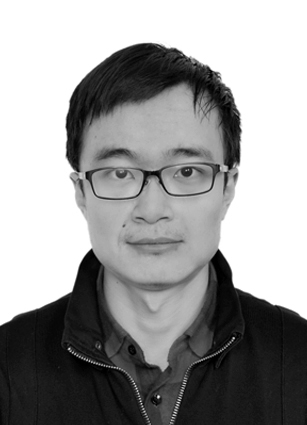}}]{Donghao Xu}
received B.S. degree in information and computing science in 2012
from Peking University, China. In 2018, he obtained Ph.D. degree in computer science from the same university. He is currently a post-doctoral researcher in computer science in the Key Lab of Machine Perception (MOE), Peking University, China. His research interests include computer vision, machine learning and intelligent vehicles.
\end{IEEEbiography}

\begin{IEEEbiography}[{\includegraphics[width=1in,height=1.25in,clip,keepaspectratio]{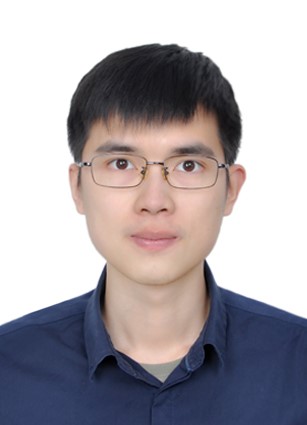}}]{Zhezhang Ding}
received the B.S. degree in computer science (intelligent science and
technology) from Peking University, Beijing, China,
in 2018. He is currently pursuing the M.S. degree in
intelligent robots with the Key Laboratory of Machine
Perception, Peking University, Beijing, China
His research interests include intelligent vehicles and
machine learning.
\end{IEEEbiography}

\begin{IEEEbiography}[{\includegraphics[width=1in,height=1.25in,clip,keepaspectratio]{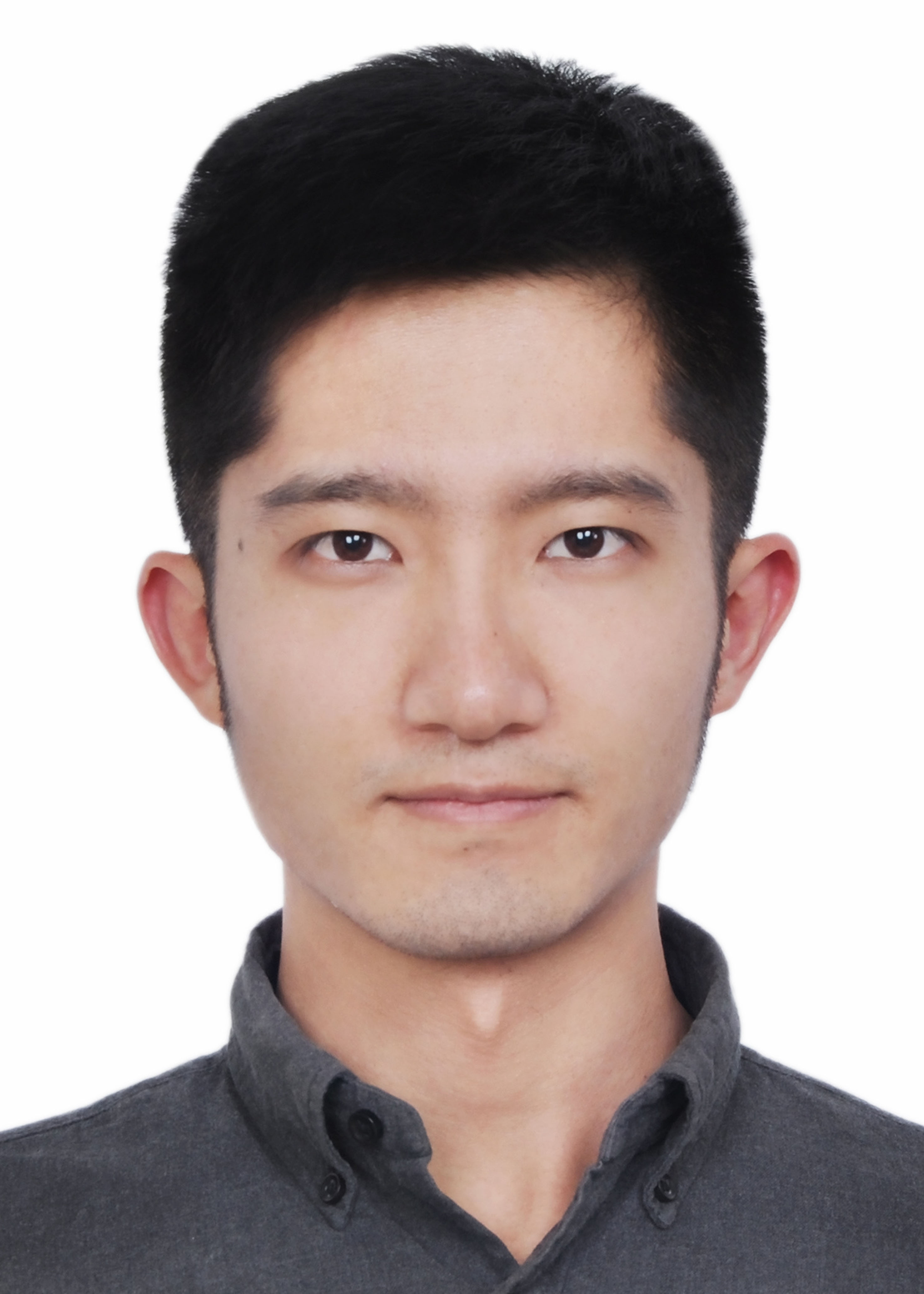}}]{Chenfeng Tu}
received B.S. degree in Electronic Science and Technology from Southeast University, Nanjing, China in 2015 and a M.S. degree in Microelectronics from Peking University, Beijing, China in 2018. He was a research assistant with the Key Laboratory of Machine Perception, Peking University from 2017 to 2018. He is currently pursuing a M.S. degree in robotics in Carnegie Mellon University, USA. His research interests include intelligent vehicles, SLAM and sensor calibration.
\end{IEEEbiography}

\begin{IEEEbiography}[{\includegraphics[width=1in,height=1.25in,clip,keepaspectratio]{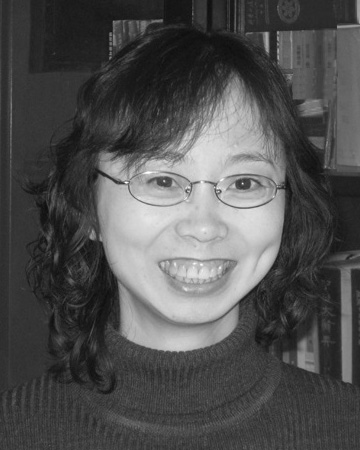}}]{Huijing Zhao}
received B.S. degree in computer science in 1991 from
Peking University, China. From 1991 to 1994, she was recruited by Peking
University in a project of developing a GIS platform. She obtained M.E.
degree in 1996 and Ph.D. degree in 1999 in civil engineering from the
University of Tokyo, Japan. After post-doctoral research
as the same university, in 2003, she was promoted to be a visiting
associate professor in Center for Spatial Information Science, the
University of Tokyo, Japan. In 2007, she joined Peking Univ as an associate
professor at the School of Electronics Engineering and Computer Science.
Her research interest covers intelligent vehicle, machine perception and mobile robot.
\end{IEEEbiography}

\begin{IEEEbiography}[{\includegraphics[width=1in,height=1.25in,clip,keepaspectratio]{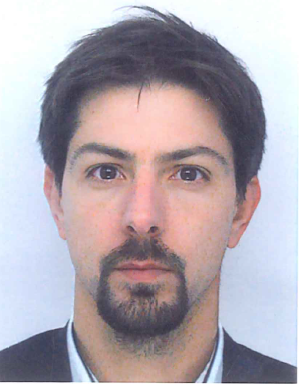}}]{Mathieu Moze}
 was born in Bordeaux, France, in 1979. He received a M.Eng. degree in Mechatronics from French Ecole Nationale d'Ingenieurs (ENIT) and a M.S. degree in Control Systems from Institut National Polytechnique Toulouse (INPT),  both  in  2003, before a Ph.D. degree in Control Theory from Bordeaux University in 2007.
In 2008, he became a consultant for industrial firms and was associated with IMS Laboratory in Bordeaux, where he studied algebraic approaches to fractional order systems analysis and robust control theory.
Since 2010, he  has been with the Scientific Department of Groupe PSA where he conducts advanced research concerning modeling, design and control of mechatronic systems for automotive applications, mainly Autonomous Driving.

\end{IEEEbiography}

\begin{IEEEbiography}[{\includegraphics[width=1in,height=1.25in,clip,keepaspectratio]{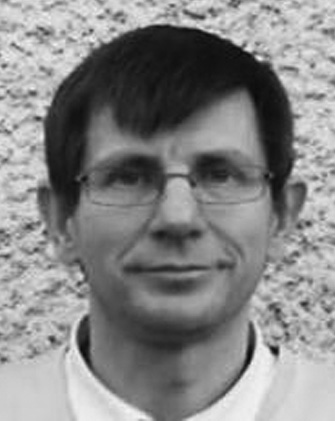}}]{Fran\c{c}ois Aioun}
received an engineering degree in electronics, computer science and Automatic control in 1988 from
ESIEA high school, France. In 1989, he obtained a post-graduate diploma in Automatic control and Signal processing.
From 1989 to 1993, she was recruited by Electricit\'e de France to study active vibration control of a structure.
He obtained a Ph.D. degree in Automatic Control in 1993.
After post-doctoral research at Ecole Normale sup\'erieure of Cachan and in different companies, he joined Groupe PSA in 1997.
His research interest in automatic control covers active vibration, power plant, powertrain, actuators and more recently Autonomous
and Intelligent vehicles.
\end{IEEEbiography}

\begin{IEEEbiography}[{\includegraphics[width=1in,height=1.25in,clip,keepaspectratio]{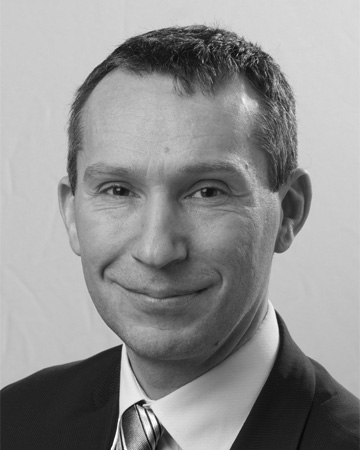}}]{Franck Guillemard}
was born in France in March 18th, 1968. He obtained his Ph.D. degree in Control Engineering from the University of Lille, France, in 1996. Presently he works in the Scientific Department of Groupe PSA where he is in charge of advanced research concerning Computing Science, Electronics, Photonics and Control. He is also expert in the field of modeling, design and control of mechatronic systems for automotive.
\end{IEEEbiography}

\end{document}